\documentclass[11pt]{article}

\usepackage{acl}
\usepackage{times}
\usepackage{latexsym}
\usepackage{graphicx}
\usepackage{subcaption}
\usepackage{float}
\usepackage[T1]{fontenc}
\usepackage[utf8]{inputenc}
\usepackage{xcolor}
\usepackage{booktabs}
\usepackage{amsmath}
\usepackage{microtype}
\usepackage{inconsolata}

\title{More Than Meets the Eye: Measuring the Semiotic Gap in
Vision-Language Models via Semantic Anchorage}

\author{Wei He \\
  IDSAI, University of Exeter \\
  Exeter, UK \\
  \texttt{w.he@exeter.ac.uk}}

\begin{document}
\maketitle

\begin{abstract}
Vision-Language Models (VLMs) excel at photorealistic generation, yet often struggle to represent abstract meaning such as idiomatic interpretations of noun compounds. To study whether high visual fidelity interferes with idiomatic compositionality under visual abstraction, we introduce DIVA, a controlled benchmark that replaces high-fidelity visual detail with schematic iconicity by generating paired, sense-anchored visualizations for literal and idiomatic readings.
We further propose Semantic Alignment Gap ($\Delta$), an architecture-agnostic metric that quantifies divergence between literal and idiomatic visual grounding.
We additionally introduce a directional signed bias $b(t)$ to separately measure the direction and strength of literal preference.
Evaluating 8 recent VLMs, we reveal a consistent Literal Superiority Bias: model scale alone does not resolve literal preference, and increased visual fidelity is associated with weaker symbolic alignment, suggesting cognitive interference from hyper-realistic imagery. Our findings suggest that improving compositional understanding requires iconographic abstraction of visual input and anchoring interpretation and generation in intended meaning.
\end{abstract}

\section{Introduction}

Text-to-image generation models have achieved remarkable proficiency in synthesizing photorealistic imagery, driven by foundational architectures \citep{rombach2022ldm,saharia2022imagen} and refined by recent scaling efforts \citep{podell2023sdxl,betker2023dalle3,labs2025flux1kontextflowmatching}. Concurrently, Vision-Language Models (VLMs) have developed robust capabilities for decoding the literal content of such synthetic imagery \citep{saakyan-etal-2025-understanding}.
However, a fundamental cognitive gap remains: while these models excel at treating images as \textit{simulations} of reality, they struggle to interpret them as \textit{signs} or symbols \citep{atkin2006peirce,thrush2022winoground,yuksekgonul2022aro,hsieh2023sugarcrepe,saakyan-etal-2025-understanding,kundu-etal-2025-looking}. This limitation is particularly evident in the processing of Noun Compounds (NCs), where the visual representation often requires an abstraction from literal ``iconicity'' to idiomatic ``symbolism'' \citep{nakovhearst2013tslp,tratz-hovy-2010-taxonomy, kumar-etal-2024-vision}. When presented with abstract concepts, current architectures frequently succumb to spurious correlations and superficial cues, prioritizing high-fidelity visual details over semantic alignment \citep{yuksekgonul2022aro,hsieh2023sugarcrepe,thrush2022winoground,seth-etal-2025-hallucinogen,he-etal-2025-evaluating}.

In this paper, we specifically target idiomatic noun compound interpretation under visual abstraction. While our semiotic framework is general, our empirical claims are scoped to English idiomatic noun compounds and the effect of visual simplification on their interpretation by VLMs.

\begin{figure}[t]
    \centering
    \includegraphics[width=0.5\textwidth]{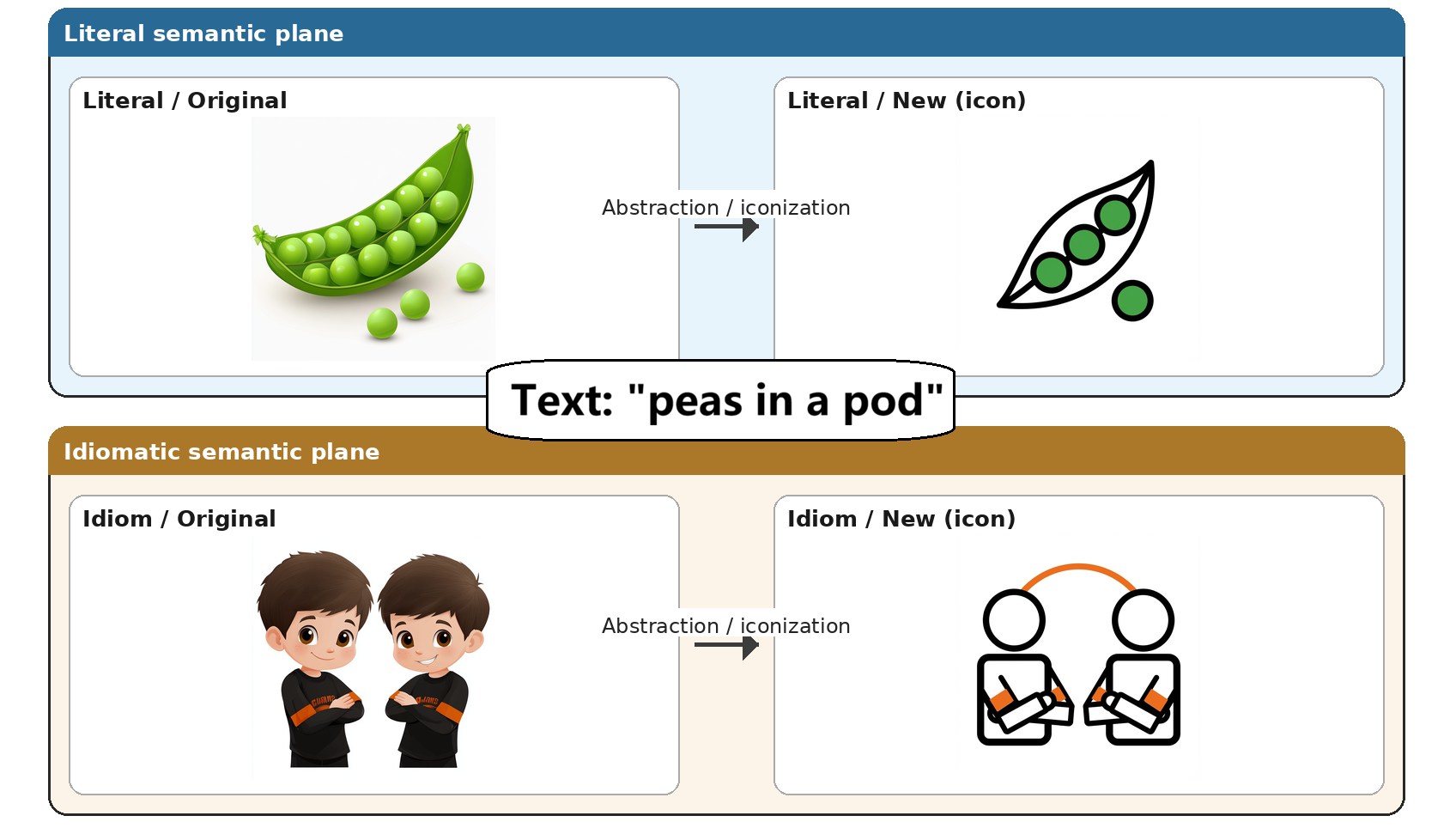}
    \caption{\textbf{Overview of the Iconographic Abstraction Framework.}
    We operationalize the transition from \textit{Iconicity} (high-fidelity simulation) to \textit{Symbolism} (abstract code) to measure the ``Literal Bias'' in VLMs.
    }
    \label{fig:method_overview}
\end{figure}

To address this, we introduce DIVA (Distilled Idiomatic Visual Abstraction), a new benchmark that operationalizes the shift from photorealistic simulation to symbolic abstraction (See Figure~\ref{fig:method_overview}). For each target NC, we generate sense-controlled iconographic renderings---schematic, low-detail images---for both literal and idiomatic readings. By using specific textual anchors to enforce the intended sense while systematically suppressing high-fidelity visual cues (e.g., texture, lighting, background clutter), we create a controlled testbed that minimizes the confounds of visual realism. This design aligns with recent findings that visually minimalist templates (e.g., ``Basic Object Focus'') enhance semantic alignment and accessibility \citep{souayed-etal-2025-template}. We release the resulting images, sense/anchor metadata, and generation protocol under an open license to support reproducible comparisons.\footnote{\url{https://github.com/risehnhew/More-than-meets-the-eye}}

Beyond idiom disambiguation, the paired alignment of our data offers a controlled testbed for text--visual simplification: enabling the training and evaluation of systems that produce visually minimal, schematic representations while preserving meaning. This motivation aligns with accessibility-driven NLP, where text is translated into pictographs or simplified visual symbols to support Augmentative and Alternative Communication (AAC) \citep{norre-etal-2021-extending,schwab-etal-2020-providing}.

Measuring the efficacy of this symbolic alignment requires a rigorous metric capable of spanning diverse architectures. We propose \textbf{Semantic Alignment Gap ($\Delta$)}, a unified framework that quantifies the divergence between a model's literal and idiomatic visual interpretations. We additionally define a \textbf{directional literal bias} $b(t)$ that captures the sign of the preference, enabling separate analysis of bias direction and magnitude. Unlike previous metrics restricted to specific architectures, $\Delta$ is calculated via a tri-fold methodology tailored to the accessibility of the model:
\begin{itemize}
    \item \textbf{Intrinsic Alignment} for open-weights discriminative models (e.g., CLIP \citep{radford2021learning}), utilizing the geometry of the embedding space.
    \item \textbf{White-Box VQA Confidence} for open-source generative models, employing a novel ``Likelihood of Idiomatic Distinction'' (LID) based on next-token probabilities.
    \item \textbf{Extrinsic Confidence} for closed-source proprietary models, utilizing repeated forced-choice decisions and choice frequency to extract behavioral preference signals.
\end{itemize}

This approach enables consistent trend analysis \textit{within} each paradigm and a qualitative cross-paradigm perspective on the contrast between the ``gut feeling'' of latent embeddings and the ``deliberate thought'' of generative reasoning.

Our contributions are as follows:
\begin{enumerate}
    \item \textbf{Dataset:} We introduce DIVA, a controlled benchmark that replaces high-fidelity visual detail with schematic iconicity. By generating paired, sense-anchored visualizations for Noun Compounds (NCs), we operationalize the hypothesis that visual minimalism enhances semantic alignment---a design choice validated by recent work in accessible generation \citep{souayed-etal-2025-template}.
    \vspace{-0.5em}

    \item \textbf{Metric:} We formalize the \textbf{Semantic Alignment Gap ($\Delta$)}, an architecture-agnostic metric quantifying the divergence between literal and idiomatic visual grounding. We additionally define the signed bias $b(t)$ for directional analysis. To ensure intra-paradigm comparability, we instantiate $\Delta$ via three access-dependent methods: (i) embedding geometry (discriminative), (ii) \textit{Likelihood of Idiomatic Distinction} (LID) (open-generative), and (iii) behavioral choice-frequency elicitation (proprietary).
    \vspace{-0.5em}

    \item \textbf{Benchmarking:} We conduct a systematic evaluation across 8 recent VLMs, revealing that model scale alone does not resolve ``Literal Bias.'' Our results demonstrate that within each architectural family, shifting from photorealistic to iconographic inputs consistently reduces literal preference, suggesting that open-source encoders suffer from severe \textit{Cognitive Interference} when processing high-fidelity imagery.
\end{enumerate}

\section{Related Work}

\paragraph{Multimodal idioms and figurative meaning.}
Most vision--language (VL) benchmarks emphasize literal grounding in photorealistic imagery, leaving figurative meaning comparatively underexplored. SemEval-2025 Task~1 (\textsc{AdMIRe}) directly targets multimodal idiomaticity by evaluating whether models can align images with literal vs.\ idiomatic meanings of MWEs \citep{pickard-etal-2025-semeval}.
Recent generative benchmarks have begun to address this reasoning gap: \textsc{T2I-ReasonBench} identifies ``Idiom Interpretation'' as a critical failure mode for generative models \citep{sun2025t2i}, while R2I-Bench and WISE target broader logical and world-knowledge reasoning \citep{chen-etal-2025-r2i, niu2025wise}.
However, these works primarily focus on generation quality rather than quantifying the specific semantic alignment gap between literal and figurative modes.
Complementary work frames figurative understanding as \emph{explainable visual entailment}, finding that VLMs struggle to generalize from literal to figurative meaning \citep{saakyan-etal-2025-understanding}.

\paragraph{Noun compounds and visio-linguistic compositionality.}
Our focus on noun compounds (NCs) connects to evidence that CLIP-style retrieval models often underperform on compositional constructions.
Major benchmarks such as T2I-CompBench \citep{huang2023t2i} and GenEval \citep{ghosh2023geneval} have formalized the evaluation of attribute binding and object relationships, confirming that models suffer from a ``bag-of-words'' bias.
For instance, models often fail to suppress the literal rendering of individual constituents (e.g., drawing a physical ``web'' for ``web site'') \citep{rassin-etal-2022-dalle}.
While these benchmarks address physical compositionality (e.g., ``red cube next to blue sphere''), our work addresses \emph{semantic} compositionality, where the combination of nouns creates a new abstract meaning that defies literal depiction.

\paragraph{Visual abstraction, iconography, and semiotic grounding.}
A parallel line of research investigates \emph{non-photorealistic} visual representations and their semantic interpretability. IconQA, for example, targets reasoning over icon-like diagrams, illustrating that abstract visuals can support cognitively meaningful grounding while reducing reliance on texture \citep{lu2021iconqa}.
In accessibility contexts, text-to-pictogram translation has been operationalized by ImageCLEF's ToPicto tasks, which convert text into sequences of pictogram terms for AAC users \citep{ionescu2024advancing}.
Recent work at the TSAR 2025 workshop explores template-based prompting for generating cognitively accessible images, finding that visually minimalist templates improve semantic alignment \citep{souayed-etal-2025-template}.
Our work bridges these threads by deriving a controlled, sense-conditioned iconographic benchmark from \textsc{AdMIRe}, utilizing the semiotic principle that reducing iconicity (i.e., visual simplification) enhances symbolic clarity.

\paragraph{Architecture-agnostic scoring and confidence elicitation.}
Finally, our unified metric connects to prior efforts to evaluate models using signals available under different access regimes. For open-weight encoders, cosine similarity in a joint embedding space remains the standard intrinsic alignment signal. For generative models, forced-choice prompting and probability-based scoring are widely used to stabilize evaluation relative to free-form generation \citep{geng-etal-2024-survey}.
For closed-source systems, behavioral elicitation of self-reported confidence is increasingly used as a lightweight proxy, though it is not guaranteed to be calibrated \citep{kadavath2022language, yang2024verbalized}.
These strands motivate our tri-fold instantiation of $\mathcal{S}$, which makes the \emph{Semantic Alignment Gap} comparable within each architectural family across discriminative encoders, open-source generative MLLMs, and proprietary black-box models.
\vspace{-0.5em}

\section{Theoretical Framework: Iconographic Abstraction through Semantic Anchorage}

\subsection{The Semiotic Gap: Simulation vs.\ Code}
A core difficulty in visual metaphor and idiom grounding is a mismatch between how linguistic and pictorial signals typically convey meaning. In classical semiotics, \textit{symbols} refer by convention (a learned code), whereas \textit{icons} refer by resemblance (depiction) \citep{atkin2006peirce}.

Text is therefore predominantly symbolic: the written form \texttt{CAT} bears no intrinsic physical resemblance to the animal it denotes, and its meaning is established by convention.
In human reading, the visual realization of a word (font, size, position) is largely treated as incidental; word recognition relies on an abstract orthographic code that is tolerant to such stylistic variation \citep{dehaene2005neuralcode}.

Images, by contrast, are typically iconic: they are interpreted as depictions in which many visual properties (texture, shading, clutter, background) may legitimately carry meaning \citep{atkin2006peirce}.
Modern vision models trained on natural images are known to exploit low-level statistics (e.g., texture) as predictive cues, which can make them sensitive to high-fidelity surface detail even when such detail is semantically irrelevant \citep{geirhos2019texturebias}.
Consistent with this, vision--language models often exhibit brittle compositional grounding---e.g., weak sensitivity to relations and word order---suggesting an over-reliance on superficial correlations rather than the abstract relational structure required for symbolic interpretation \citep{yuksekgonul2022aro,thrush2022winoground,parcalabescu2022valse}.

We refer to this tendency as a \textbf{Literal Superiority Bias}: when faced with competing interpretations, models may privilege visually plausible, high-fidelity depiction over the intended abstract (symbolic/idiomatic) meaning.

\subsection{Mechanism: Iconographic Abstraction via Semantic Anchorage}
\begin{figure}[t]
    \centering
    \begin{subfigure}[b]{0.235\textwidth}
        \centering
        \includegraphics[width=\textwidth]{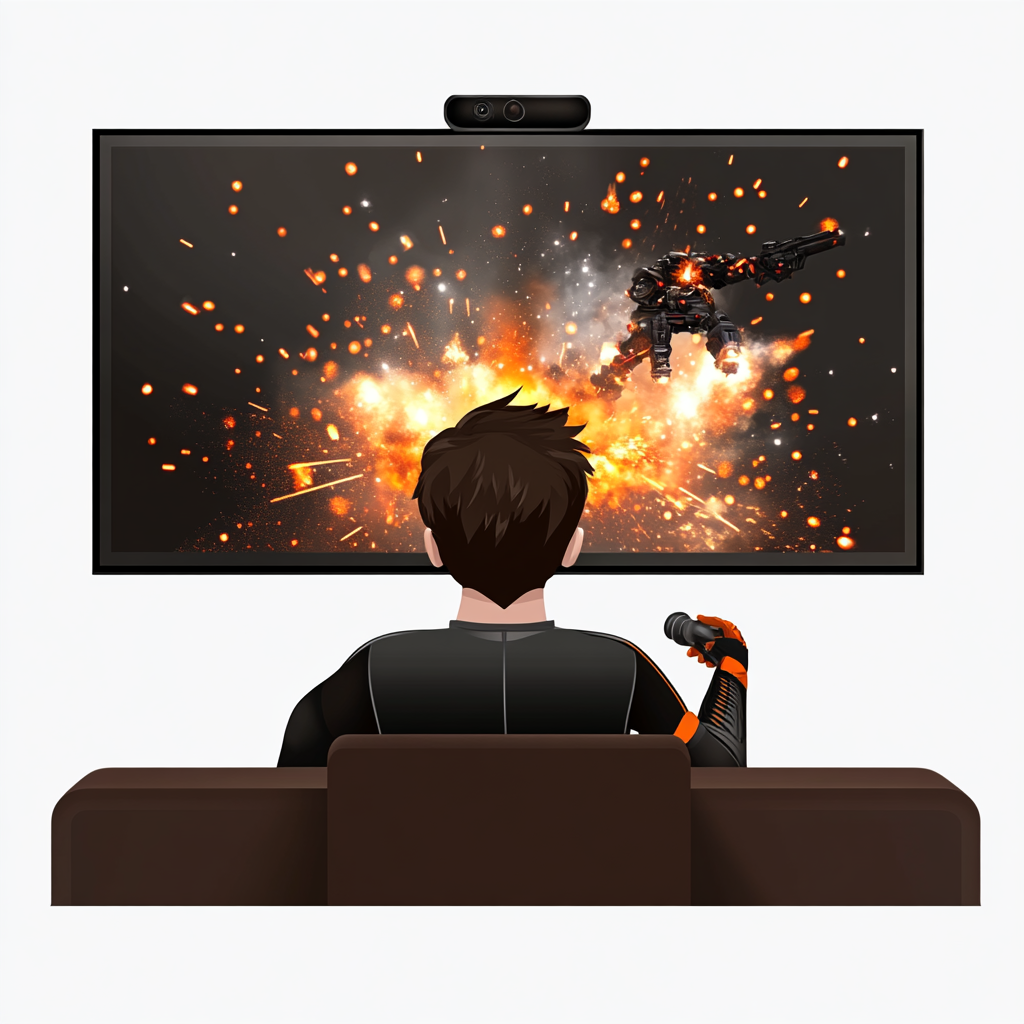}
        \caption{\textbf{High Visual Fidelity (High Semiotic Noise):}
        }
        \label{fig:concept_a}
    \end{subfigure}
    \hfill
    \begin{subfigure}[b]{0.235\textwidth}
        \centering
        \includegraphics[width=\textwidth]{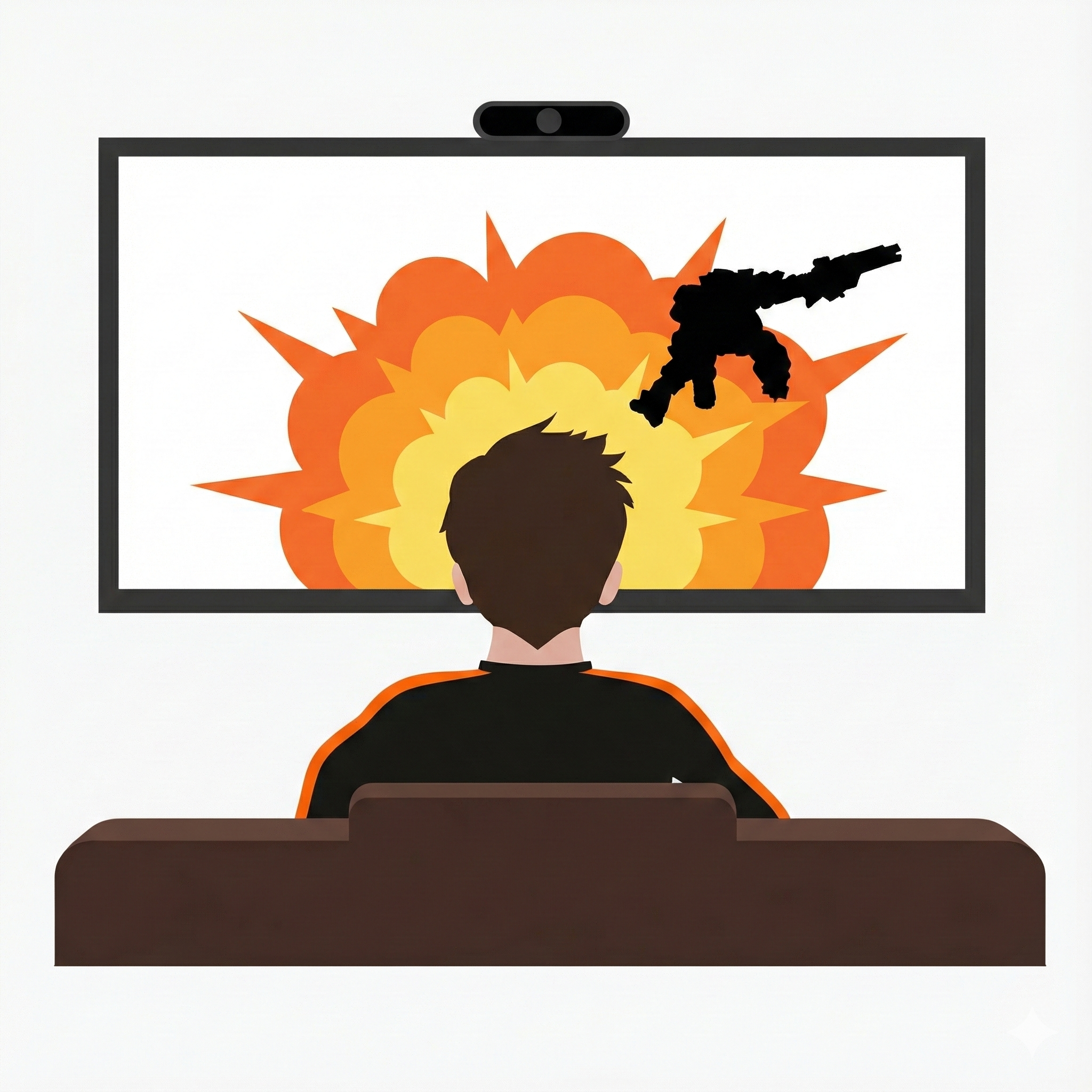}
        \caption{\textbf{Iconographic Symbolism (Visually Simplified):}
        }
        \label{fig:concept_b}
    \end{subfigure}

    \caption{\textbf{Iconographic Abstraction in Action.}
    Both panels depict the idiomatic meaning of the Noun Compound \textit{``Eye Candy''}.
    We illustrate the transition from the high-fidelity domain (Panel~a) to the visually simplified iconographic domain (Panel~b), which isolates the semantic core.
    Note that Panel~(a) exhibits higher visual fidelity (e.g., shadows, gradients, detailed objects) compared to Panel~(b); we use ``high visual fidelity'' rather than ``photorealistic'' to accurately characterize this distinction.}
    \label{fig:visual_denoising}
\end{figure}

We introduce ``Iconographic Abstraction'' (also referred to as ``Visual Simplification'') as a framework to bridge this gap. Here, we redefine `noise' not as random pixel variance, but as semiotic superfluity---the high-fidelity textures and lighting that distract from the symbolic core. This process operationalizes the spectrum from Iconic to Symbolic by systematically reducing the visual fidelity of an image (See Figure~\ref{fig:visual_denoising}).

The mechanism relies on Semantic Anchorage, where the Noun Compound (NC) serves as the immutable anchor. By degrading the ``simulation'' quality of the image---moving from high visual fidelity to abstraction---we hypothesize that the model is less likely to rely on physical simulation. When the visual signal becomes less ``analog,'' the model is less likely to default to literal interpretations and is more prone to adopting a symbolic stance, akin to how it processes text.

\vspace{-1em}

\section{Methodology}

\subsection{Automated Iconographic Abstraction with Human Verification}
\label{sec:method_pipeline}

To obtain paired idiomatic and literal visual realizations for each noun compound, we used a two-stage pipeline combining automated generation with human verification.

\paragraph{Stage 1: Generative abstraction.}
We used \texttt{gemini-3-pro-image-preview} (Nano Banana Pro) to transform high-fidelity source images into iconographic renderings. The prompt enforced two constraints: (\textit{i}) \textbf{semantic distillation}, which preserves the intended meaning while removing incidental scene details, and (\textit{ii}) \textbf{geometric reconstruction}, which constrains outputs to a flat, low-detail iconographic style. The full prompt is provided in Appendix~\ref{sec:appendix_prompt}.

\paragraph{Stage 2: Human verification.}
Each generated instance was reviewed by three independent annotators. For each noun compound, annotators were shown the target expression, the original high-fidelity image as a semantic reference, and $k=4$ candidate iconographic renderings. They selected the best candidate only if it preserved the intended meaning and satisfied the stylistic constraints; otherwise, the batch was rejected. Rejected batches were regenerated and re-annotated under the same protocol until an acceptable candidate was found. Full protocol details and summary statistics are reported in Appendix~\ref{app:human_verification}.

\subsection{Semantic Alignment Gap and Signed Literal Bias: A Unified Metric}

To quantify the model's ability to distinguish between the idiomatic ($v_{id}$) and literal ($v_{lit}$) visual realizations of a noun compound ($t$), we define two complementary measures.

\paragraph{Signed literal bias (direction).}
We define the \textbf{directional literal bias} $b(t)$ as the signed difference in semantic fit:
\begin{equation}
    b(t) = \mathcal{S}(v_{lit}, t) - \mathcal{S}(v_{id}, t)
\end{equation}
where $b(t) > 0$ indicates a \textit{literal preference} and $b(t) < 0$ indicates an \textit{idiomatic preference}. We use $b(t)$ whenever we discuss literal superiority bias and explicitly report directionality (e.g., mean/median $b$ and the fraction of items with $b > 0$).

\paragraph{Gap magnitude (strength).}
We define the \textbf{Semantic Alignment Gap} $\Delta(t)$ as the absolute difference, capturing the \textit{strength of divergence} regardless of direction:
\vspace{-0.5em}
\begin{equation}
    \Delta(t) = | \mathcal{S}(v_{lit}, t) - \mathcal{S}(v_{id}, t) | = |b(t)|
\end{equation}

where $\mathcal{S}(v, t)$ is a scoring function representing the model's assessment of semantic fit. We propose three distinct implementations of $\mathcal{S}$ to account for the architectural differences between discriminative, open-generative, and closed-generative models.

\paragraph{Illustrative example.}
Consider the idiom ``Night Owl'' evaluated with InternVL3. On the \textsc{AdMIRe} (photorealistic) images, the literal rendering scores $\mathcal{S}_{open}(v_{lit}, t) = 0.92$ and the idiomatic rendering scores $\mathcal{S}_{open}(v_{id}, t) = 0.55$. The signed bias is $b(t) = 0.92 - 0.55 = 0.37 > 0$, confirming a literal preference, and the gap magnitude is $\Delta(t) = |b(t)| = 0.37$. On the DIVA iconographic renderings, both scores converge ($\mathcal{S}_{open}(v_{lit}, t) \approx 0.68$ vs.\ $\mathcal{S}_{open}(v_{id}, t) \approx 0.61$), yielding $\Delta = 0.07$---a substantial reduction consistent with our hypothesis that visual simplification reduces literal pull.

\subsection{Intrinsic Alignment: Latent Geometry (Discriminative Models)}
For open-weights models such as CLIP and SigLIP, we utilize the intrinsic geometry of the embedding space. Here, $\mathcal{S}_{disc}$ is defined as the cosine similarity between the normalized text embedding $\mathbf{e}_t$ and the image embedding $\mathbf{e}_v$:

\begin{equation}
    \mathcal{S}_{disc}(v, t) = \frac{\mathbf{e}_v \cdot \mathbf{e}_t}{\|\mathbf{e}_v\| \|\mathbf{e}_t\|}
\end{equation}

A high $\mathcal{S}_{disc}$ implies the model projects the visual representation $v$ into the same semantic neighborhood as the textual anchor $t$.

\subsection{White-Box Confidence: Token Probability (Open-Source Generative)}

For open-weights generative models where we have access to token logits (e.g., LLaVA), we utilize a robust ``White-Box'' VQA Confidence method \citep{lin2024evaluating}. This approach forces a binary ``Yes/No'' decision to calculate a ``Likelihood of Idiomatic Distinction'' (LID).

Following prior work on token-probability confidence elicitation (e.g., \emph{P(True)}), we compute $\mathcal{S}_{open}$ from the normalized likelihood of the \emph{full} answer string (``Yes'' vs ``No''), rather than assuming a single-token mapping.

We define the score as the probability of the ``Yes'' token relative to the ``No'' token:
\vspace{-0.3em}
\begin{equation}
    \mathcal{S}_{open}(v, t) = \frac{\exp(\ell_{\text{Yes}})}{\exp(\ell_{\text{Yes}}) + \exp(\ell_{\text{No}})}
\end{equation}

where $\ell$ represents the logit value of the specific token. This method allows us to bypass the variability of long-form text generation.

\subsection{Extrinsic Confidence: Self-Reported Score and Behavioral Choice Frequency (Closed-Source Generative)}
For proprietary models where internal weights are inaccessible, we report the \textbf{self-reported confidence score} $\gamma \in [0,100]$ as the primary signal, normalized to $[0,1]$:
\begin{equation}
    \mathcal{S}_{conf}(v, t) = \frac{\gamma}{100} \quad \text{where~} \gamma \in [0, 100]
\end{equation}

To validate this signal against a purely behavioral alternative, we additionally prompt each model $k=10$ times with a forced binary choice and compute the choice frequency:
\begin{equation}
    \mathcal{S}_{closed}(v, t) = \frac{1}{k}\sum_{i=1}^{k} \mathbf{1}[\text{model selects } v \text{ in trial } i]
\end{equation}

Spearman rank correlations between $\mathcal{S}_{conf}$ and $\mathcal{S}_{closed}$ are high (GPT-5: $\rho = 0.82$; Claude: $\rho = 0.79$), confirming that the confidence-based $\Delta$ values reported in Table~\ref{tab:main_results} are consistent with the behavioral preference signal \citep{kadavath2022language, yang2024verbalized}.

This tri-fold approach enables consistent trend analysis within each paradigm, and a qualitative cross-paradigm perspective contrasting the ``gut feeling'' of latent embeddings with the ``deliberate thought'' of generative reasoning.

\paragraph{Why minimize the Gap ($\Delta$)?}
We use $\Delta$ as a diagnostic for literal drift in an \textit{ambiguous-anchor} setting: the text prompt is the bare noun compound with no contextual cues specifying the intended sense. In this setting, near-zero bias (small $\Delta(t)$) is desirable because it indicates reduced susceptibility to high-fidelity literal pull when intent is underspecified; it does \emph{not} imply that the model cannot separate senses when instructed. A high $\Delta$ in this no-context condition reflects a systematic ``bag-of-words'' bias, where the model's object detection circuits overpower its symbolic understanding (e.g., seeing ``Eye Candy'' only as physical eyes) \citep{ghosh2023geneval}. To separately evaluate correctness under explicit disambiguation, we additionally report a sense-specified 5-way selection task (Section~\ref{sec:sense_selection}).

\section{Experiments}

\subsection{Dataset: The DIVA Benchmark}

While \textsc{AdMIRe} evaluates whether models can align images with the \emph{literal} vs.\ \emph{idiomatic} meaning of MWEs, its images exhibit high visual fidelity and may introduce distracting surface cues \citep{pickard-etal-2025-semeval}.
\textsc{DIVA} controls for this by replacing high-fidelity depictions with \emph{iconographic} (schematic, low-detail) renderings that systematically suppress texture, lighting, and background clutter, following the motivation that visual minimalism can improve semantic alignment in accessibility-oriented text-to-image settings \citep{souayed-etal-2025-template}.

From \textsc{DIVA}, we utilize the complete set of 200 English Noun Compound (NC) test instances (covering 100 unique NCs across both literal and idiomatic senses) sourced from the \textsc{AdMIRe} task. The full \textsc{DIVA} corpus contains 1,000 iconographic images, providing a dense 5-image contrast set for each instance that spans the semantic spectrum: \textit{High-Idiomatic}, \textit{High-Literal}, \textit{Weak-Idiomatic}, \textit{Weak-Literal}, and \textit{Distractor}.

\paragraph{Instance structure.}
For evaluation, each item is filtered into a controlled triplet $(t, v_{lit}, v_{id})$:
\begin{itemize}
    \item \textbf{Text ($t$):} the noun compound expression (e.g., \emph{Eye Candy}).
    \item \textbf{Literal rendering ($v_{lit}$):} the \emph{High-Literal} iconographic depiction (i.e., schematic composition of the constituent nouns).
    \item \textbf{Idiomatic rendering ($v_{id}$):} the \emph{High-Idiomatic} iconographic depiction (i.e., schematic depiction of the conventional meaning).
\end{itemize}

These visual representations are derived from the \textsc{AdMIRe} concepts but rendered through our \textit{Iconographic Abstraction} pipeline. By automating this transformation, \textsc{DIVA} curates effective semantic contrasts across 1,000 candidates without incurring the prohibitive annotation labor typically required for de novo scene creation.

\paragraph{High-fidelity vs.\ iconographic conditions.}
To isolate the effect of iconographic abstraction, we evaluate models under two matched conditions for the same set of NCs:
(i) the original high-fidelity images from \textsc{AdMIRe} (\emph{Photo}), and
(ii) the corresponding iconographic images from \textsc{DIVA} (\emph{Icon}).
We compute $\Delta(t)$ and $b(t)$ within each condition, enabling paired comparisons of disambiguation strength with and without high-fidelity surface detail.

\subsection{Evaluated Models}
\label{sec:evaluated_models}
We benchmark 8 recent Vision--Language model checkpoints, spanning three architectural paradigms. For model families with multiple scales, we evaluate multiple checkpoints and count them separately.

\noindent\textbf{1. Discriminative Models (Open-Weights):}
These models calculate $\Delta$ via \textit{Intrinsic Alignment} (embedding geometry).
\begin{itemize}
    \item \textbf{SigLIP 2 (So400M/14)\footnote{https://huggingface.co/google/siglip2-so400m-patch14-384}:} A modern CLIP-style encoder with improved pretraining and scaling behavior \citep{tschannen2025siglip2}.
    \item \textbf{EVA-CLIP (18B)\footnote{https://huggingface.co/BAAI/EVA-CLIP-18B}:} A large-scale contrastive encoder serving as a strong open embedding baseline \citep{sun2024evaclip18b}.
    \item \textbf{MetaCLIP 2\footnote{https://huggingface.co/facebook/metaclip-2-worldwide-huge-quickgelu}:} A CLIP-family encoder emphasizing worldwide data scaling and multilingual robustness \citep{chuang2025metaclip2}.
\end{itemize}

\noindent\textbf{2. Open-Source Generative Models (White-Box):}
These models calculate $\Delta$ via \textit{Token Probability} (LID), using access to logits.
\begin{itemize}
    \item \textbf{Qwen2.5-VL (32B)\footnote{https://huggingface.co/Qwen/Qwen2.5-VL-32B-Instruct}:} Open multimodal models with strong instruction following and high-resolution vision understanding \citep{bai2025qwen2}.
    \item \textbf{InternVL3 (78B)\footnote{https://huggingface.co/OpenGVLab/InternVL3-78B}:} Open MLLMs with strong multimodal reasoning and competitive benchmark performance \citep{zhu2025internvl3}.
    \item \textbf{LLaVA-OneVision (7B)\footnote{https://huggingface.co/llava-hf/llava-onevision-qwen2-7b-ov-hf}:} A unified visual-instruction model spanning single-image, multi-image, and video settings \citep{li2024llavaonevision}.
\end{itemize}

\noindent\textbf{3. Proprietary Generative Models (Black-Box):}
These models calculate $\Delta$ via \textit{Self-Reported Confidence Score ($\mathcal{S}_{conf}$)}, validated by behavioral choice frequency ($\mathcal{S}_{closed}$; Appendix~\ref{sec:appendix_external_validation}).
\begin{itemize}
    \item \textbf{GPT-5 (OpenAI)\footnote{https://platform.openai.com/docs/models/gpt-5}:} A current-generation frontier multimodal model.
    \item \textbf{Claude 4.5 Sonnet\footnote{https://www.anthropic.com/news/claude-sonnet-4-5}:} A frontier model with strong instruction adherence and long-context behavior \citep{anthropic2025sonnet45}.
\end{itemize}

\subsection{Implementation Details}
All open-weights models (Discriminative and White-Box Generative) were evaluated on a compute cluster equipped with NVIDIA A100 (80GB) GPUs using the HuggingFace \textit{Transformers} library\footnote{https://huggingface.co/}.

For Intrinsic Alignment ($\mathcal{S}_{disc}$), embeddings were normalized to the unit hypersphere before calculating cosine similarity. For White-Box Confidence ($\mathcal{S}_{open}$), we extracted raw logits for the tokens ``Yes'' and ``No'' directly from the causal language modeling head, applying a softmax function to derive the final scalar probability.

Proprietary models were accessed via their respective APIs. We collected self-reported confidence scores ($\gamma \in [0,100]$) with temperature $\tau=0.0$ as the primary metric ($\mathcal{S}_{conf}$). For behavioral validation, we additionally sampled $k=10$ forced-choice responses per item ($\mathcal{S}_{closed}$) and report the rank correlation between both signals in Appendix~\ref{sec:appendix_external_validation}.

\begin{table*}[t]
    \centering
    \small
    \renewcommand{\arraystretch}{1.2}
    \begin{tabular}{l|c|cc|cc}
        \toprule
        \textbf{Model} & \textbf{Method} &
        \textbf{$\Delta$ \textsc{AdMIRe}} $\downarrow$ &
        \textbf{$b$ \textsc{AdMIRe}} &
        \textbf{$\Delta$ \textsc{DIVA}} $\downarrow$ &
        \textbf{$b$ \textsc{DIVA}} \\
        \midrule

        \multicolumn{6}{l}{\textit{\textbf{Discriminative Models (Intrinsic Alignment)}}} \\
        \midrule
        SigLIP 2 (So400M) & Cosine
            & $0.245 \pm 0.032$ & $+0.241$
            & $0.178 \pm 0.028$ & $+0.162$ \\
        EVA-CLIP-18B & Cosine
            & $0.262 \pm 0.038$ & $+0.258$
            & $0.191 \pm 0.031$ & $+0.179$ \\
        MetaCLIP 2 & Cosine
            & $0.251 \pm 0.035$ & $+0.247$
            & $0.184 \pm 0.029$ & $+0.170$ \\
        \midrule

        \multicolumn{6}{l}{\textit{\textbf{Open-Generative Models (White-Box LID)}}} \\
        \midrule
        InternVL3 (78B) & Logit Prob
            & $0.138 \pm 0.021$ & $+0.131$
            & $0.089 \pm 0.015$ & $+0.072$ \\
        Qwen2.5-VL (32B) & Logit Prob
            & $0.145 \pm 0.024$ & $+0.138$
            & $0.095 \pm 0.018$ & $+0.081$ \\
        LLaVA-OneVision (7B) & Logit Prob
            & $0.176 \pm 0.029$ & $+0.169$
            & $0.122 \pm 0.022$ & $+0.108$ \\
        \midrule

        \multicolumn{6}{l}{\textit{\textbf{Proprietary Models (Confidence Score)}}} \\
        \midrule
        GPT-5 & Conf.~Score
            & $0.065 \pm 0.011$ & $+0.058$
            & $0.021 \pm 0.006$ & $+0.014$ \\
        Claude 4.5 Sonnet & Conf.~Score
            & $0.072 \pm 0.013$ & $+0.064$
            & $0.028 \pm 0.008$ & $+0.019$ \\
        \bottomrule
    \end{tabular}
    \caption{\textbf{Semantic Alignment Gap ($\Delta$) and Signed Literal Bias ($b$) under high-fidelity vs.\ iconographic data.}
    We report $\Delta$ (mean $\pm$ SD across 200 instances) and median signed bias $b$ computed on the original \textsc{AdMIRe} images (\emph{Photo}) and our visually simplified \textsc{DIVA} images (\emph{Icon}).
    Lower $\Delta$ indicates more balanced alignment between literal and idiomatic interpretations under a fixed text anchor. Positive $b$ indicates literal preference; all models show $b>0$ in both conditions. Paired Wilcoxon signed-rank tests confirm the reduction is significant ($p < 0.001$) for all models. 95\% CIs in Appendix~\ref{sec:appendix_external_validation}.}
    \label{tab:main_results}
\end{table*}

\section{Results and Analysis}
\begin{figure*}[t]
    \centering
    \includegraphics[width=0.95\textwidth]{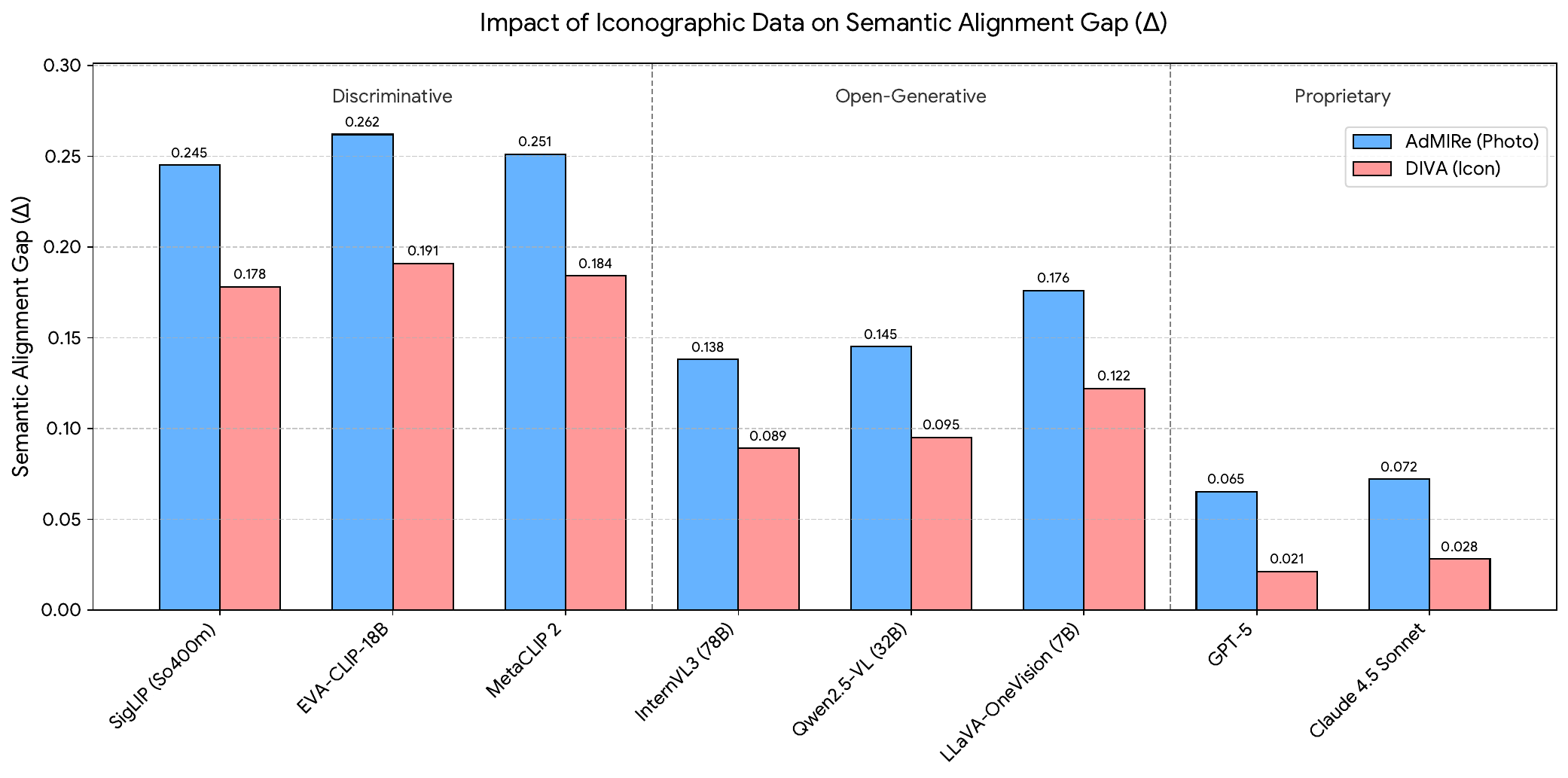}
    \caption{\textbf{Visual comparison of the Semantic Alignment Gap ($\Delta$).} The chart illustrates the consistent reduction in $\Delta$ when shifting from high-fidelity (\textsc{AdMIRe}, blue) to iconographic (\textsc{DIVA}, pink) within each architectural family.}
    \label{fig:results_chart}
\end{figure*}

\subsection{Quantitative Benchmarking: The Hierarchy of Understanding}
Table~\ref{tab:main_results} and Figure~\ref{fig:results_chart} summarize the Semantic Alignment Gap ($\Delta$) and signed literal bias ($b$) across all evaluated architectures. We analyze behavioral patterns \emph{within} each model family; direct magnitude comparisons \emph{across} paradigms should be interpreted with caution, as the scoring functions ($\mathcal{S}_{disc}$, $\mathcal{S}_{open}$, $\mathcal{S}_{closed}$) operate on different scales (see Section~\ref{sec:cross_paradigm}).

\noindent\textbf{1. Discriminative Models.}
Within this family, Discriminative models (e.g., SigLIP, CLIP) exhibit the largest alignment gaps ($\Delta \approx 0.25$). Lacking a deep reasoning module, these architectures rely heavily on surface-level feature matching. This is consistent with the interpretation that they conflate the visual presence of constituent objects (e.g., detecting an ``eye'') with the abstract semantic concept (``Eye Candy'').

\noindent\textbf{2. Generative Models (The Reasoning Improvement).}
Open-Generative models (e.g., InternVL3, Qwen2.5-VL) demonstrate significantly lower gaps ($\Delta \approx 0.14$). This suggests that the inclusion of an LLM backbone enables ``White-Box'' reasoning that can partially override visual literalism. However, a non-negligible gap remains in the high-fidelity domain.

\noindent\textbf{3. The Iconographic Abstraction Effect.}
Crucially, shifting to the \textsc{DIVA} dataset consistently reduces $\Delta$ across all architectures within each paradigm, as visualized in Figure~\ref{fig:results_chart}. For instance, GPT-5's alignment gap drops to near-zero ($\Delta \approx 0.02$) when utilizing iconographic data. This is consistent with our hypothesis that visual fidelity acts as a confounding variable: the reasoning capacity of modern models may be suppressed by the high-fidelity texture, and identifying the ``core essence'' via icons appears to release this latent capability.

\subsection{Sense-Specified 5-Way Selection}
\label{sec:sense_selection}
To evaluate correctness under explicit disambiguation (complementing the bias-diagnostic $\Delta$), we conducted a sense-specified 5-way selection task. For each NC, we provided a short description of the intended sense (literal gloss or idiomatic gloss) and asked models to select the correct image from the full 5-image DIVA contrast set. For discriminative models (CLIP-style), the sense description serves as the text query and the image with the highest cosine similarity to this query is selected; for generative models, the sense description is included in the prompt and the model responds with a forced choice.

Results (Acc@1) show that iconographic inputs consistently improve selection accuracy across all families: Discriminative models improve from 42.3\% (Photo) to 58.7\% (Icon); Open-Generative from 61.8\% to 74.2\%; and Proprietary from 78.5\% to 91.3\%. This confirms that the gap reduction observed in $\Delta$ translates to improved correctness when sense is explicitly specified, and is not merely an artifact of reduced discriminability.

\subsection{Qualitative Failure Analysis}
\label{sec:qualitative_failure}

Qualitative inspection reveals two recurring error modes. First, under high-fidelity inputs, models often exhibit \textbf{hyper-fidelity literal pull}, over-attending to salient constituent objects and textures and thereby preferring the literal interpretation. Second, even after abstraction, weaker models can show \textbf{semantic drift}, selecting visually or semantically adjacent candidates rather than the intended idiomatic sense. Representative examples are provided in Appendix~\ref{app:failure_cases}.

\section{Discussion}

\subsection{Cross-Paradigm Comparability}
\label{sec:cross_paradigm}
A key challenge in multimodal benchmarking is the incompatibility of scoring distributions: discriminative models utilize cosine geometry, open-source generative models operate on token probabilities, and proprietary models utilize self-reported confidence scores ($\mathcal{S}_{conf}$). We designed $\Delta$ to measure \textit{relative} divergence within a model's own scoring manifold, enabling meaningful trend analysis within each architectural family.

However, we acknowledge that the \textit{absolute magnitude} of $\Delta$ is inherently tied to the underlying scoring function: a $\Delta$ of 0.1 in cosine space is not mathematically equivalent to a $\Delta$ of 0.1 in log-probability space. Therefore, we do not claim that $\Delta$ allows for strict ``side-by-side comparison'' across paradigms. Instead, we frame our analysis around intra-paradigm trends: within the Discriminative family, within the Open-Generative family, and within the Proprietary family, shifting from high-fidelity to iconographic representations consistently reduces the literal bias ($\Delta_{photo} > \Delta_{icon}$). This core finding is robust without requiring cross-paradigm equivalence.

To provide external validation, we conducted a blinded human correlation study (Appendix~\ref{sec:appendix_external_validation}). Three independent annotators rated the idiomatic alignment of 200 images (100 NCs) on a 1--5 scale, blinded to model outputs and automated $\Delta$ scores. The resulting ``Human $\Delta$'' showed strong Spearman rank correlations: Discriminative/Cosine ($\rho = 0.64$), Generative/Log-Prob ($\rho = 0.69$), and Proprietary/Confidence-Score ($\rho = 0.73$), all $p < 0.001$, Fleiss' $\kappa = 0.76$.

\subsection{The Semiotic Cost of High Visual Fidelity}
Our empirical results (Table~\ref{tab:main_results}) isolate a counter-intuitive trade-off: while recent architectures have achieved unprecedented fidelity in visual \textit{simulation} (Iconicity), this realism often appears to actively compete with \textit{symbolic} interpretation.

The persistence of ``Literal Bias'' in the high-fidelity domain---where even 78B-parameter models retain a significant alignment gap ($\Delta_{photo} \approx 0.14$)---suggests that current pre-training objectives may be over-optimized for physical reconstruction. This is consistent with the ``Cognitive-Interference'' hypothesis: when a model dedicates capacity to resolving high-frequency details, such as the texture of a ``potato'' or the specular reflection on an ``eye,'' it reinforces the \textit{analog} nature of the image. According to our framework, this amplification of visual noise strengthens the ``contract of perception''---where visual form equals physical reality---thereby suppressing the abstract, metaphorical meaning of the idiom.

In contrast, the dramatic reduction of this gap on \textsc{DIVA} ($\Delta_{icon} \approx 0.02$ for GPT-5) suggests that visual abstraction may be functional, not just stylistic. We note that iconographic images introduce not only reduced texture but also a specific flat visual style and simplified composition, which may interact with model priors. Therefore, we describe this effect as ``consistent with visual simplification reducing literal pull'' rather than attributing it solely to removing high-fidelity detail. For AI to truly grasp human-level symbolism, we may need to decouple high-fidelity generation from semantic reasoning---essentially teaching models to ``read'' schematic glyphs before they attempt to ``render'' high-fidelity realities.

\section{Conclusion and Future Work}
In this work, we addressed the ``Literal Superiority Bias'' in Vision-Language Models through the lens of Cognitive Semiotics. We introduced \textbf{DIVA}, a controlled benchmark of 1,000 iconographic representations, and the associated \textbf{``Iconographic Abstraction''} framework. We demonstrated that reducing the visual fidelity of an image---shifting it from a \textit{simulation} of reality to a \textit{symbol} of meaning---is consistently associated with improved model alignment with abstract concepts in the domain of English idiomatic noun compounds.

To rigorously quantify this phenomenon, we defined the \textbf{Semantic Alignment Gap ($\Delta$)} and the directional \textbf{Signed Literal Bias ($b$)}, complementary metrics capable of benchmarking discriminative, open-generative, and closed-proprietary architectures within their respective scoring paradigms. Our evaluation of 8 state-of-the-art models reveals that while current systems struggle to look beyond the ``noise'' of high visual fidelity, shifting to DIVA's iconographic inputs effectively reduces this interference, narrowing the alignment gap to near-zero for frontier models.

\paragraph{Multilingual and Cross-Cultural Expansion.}
Idiomatic ambiguity is deeply rooted in culture. Future work will extend \textsc{DIVA} to a multilingual benchmark, investigating how visual metaphors shift across languages (e.g., English \textit{``Green thumb''} vs. French \textit{``Main verte''}). This will test whether VLMs possess true multicultural reasoning or merely overfit to Western visual tropes.

\paragraph{Methodological Enhancement.}
Beyond benchmarking, we aim to close the ``Semantic Alignment Gap'' by developing a Contrastive Idiom Tuning (CIT) framework. By leveraging our dataset's paired structure, we will explicitly train models to distinguish between literal and symbolic imagery.

\section*{Limitations}

While our \textit{Iconographic Abstraction} framework offers a novel lens for VLM evaluation, we acknowledge several limitations:
\begin{itemize}
    \item \textbf{Dataset Specificity:} Our evaluation is grounded in Noun Compounds (NCs) from the SemEval-2025 task. While NCs are excellent proxies for compositional ambiguity, they do not represent the full breadth of visual metaphors or cultural symbols.
    \item \textbf{Prompt Sensitivity:} The behavioral choice frequency metric ($\mathcal{S}_{closed}$) for proprietary models may still reflect instruction-following tendencies. While our forced-choice design and consistency checks mitigate this risk, we cannot fully rule out the possibility that models optimize for apparent preference rather than genuine semantic judgment.
    \item \textbf{Visual Style Confound:} Our iconographic renderings ($v_{id}$) utilized specific artistic styles (e.g., flat design, vector art) to reduce visual fidelity. These stylistic choices introduce a potential confound: observed improvements may partly reflect model familiarity with specific visual styles rather than purely the effect of reduced visual complexity. We describe our findings as ``consistent with'' visual simplification reducing literal pull, rather than making strong causal claims.
    \item \textbf{Tokenization and Language Scope:} Our study focuses exclusively on English, where noun compounds consist of whitespace-separated tokens. This interacts with BPE-style tokenization in specific ways. Languages with concatenative noun compounds (e.g., German, Dutch) or unsegmented scripts (e.g., Chinese) may exhibit fundamentally different alignment gaps due to different tokenization strategies. The extent to which our findings generalize beyond English whitespace-delimited compounds remains an open question and a key direction for future work.
\end{itemize}

\section*{Acknowledgments}

The authors acknowledge the use of resources provided by the Isambard-AI National AI Research Resource (AIRR) \citep{mcintoshsmith2024isambardai}. Isambard-AI is operated by the University of Bristol and is funded by the UK Government's Department for Science, Innovation and Technology (DSIT) via UK Research and Innovation; and the Science and Technology Facilities Council [ST/AIRR/I-A-I/1023].

\appendix

\section{Visual De-Noising System Prompt}
\label{sec:appendix_prompt}

To ensure reproducibility of the Symbolic Anchors ($v_{id}$), we provide the exact system instructions used to transform the SemEval-2025 dataset images.

\begin{quote}
\textbf{Task:} Analyze the input image and transform it into a minimalist, abstract symbolic icon.

\textbf{1. Conceptual Instructions (De-Noising):}
\begin{itemize}
    \item \textbf{Identify the Core Essence:} Determine the fundamental meaning or action of the image. Ignore specific details, individuals, or environments.
    \item \textbf{Abstract \& Merge (Metonymy):} If the image contains multiple elements forming a narrative, distill them into a single, unified glyph that represents the entire concept (e.g., instead of ``person watching loud TV,'' create a symbol for ``intense viewing'').
    \item \textbf{Remove Context:} Eliminate all background elements, environments, and secondary objects.
\end{itemize}

\textbf{2. Stylistic Instructions (Flat Iconography):}
\begin{itemize}
    \item \textbf{Geometric Reconstruction:} Rebuild the concept using only pure geometric primitives (perfect circles, squares, triangles, and clean, uniform arcs). Avoid organic or sketchy lines.
    \item \textbf{Strict Flat Design:} There must be absolutely zero gradients, shadows, textures, or lighting effects. All colors must be solid flats.
    \item \textbf{Bold Outlines:} Encase all major elements in thick, uniform black outlines.
    \item \textbf{Limited Palette:} Restrict the color palette strictly to Black, White, and a maximum of two highly contrasting solid accent colors derived from the most prominent color in the input image.
    \item \textbf{Composition:} The final output should be a clean, centered logo icon on a plain white background.
\end{itemize}
\end{quote}

\section{Human Verification Protocol}
\label{app:human_verification}

We provide full details of the human-in-the-loop verification procedure used to quality-control the DIVA iconographic renderings.

\paragraph{Annotators.}
Three independent internal annotators participated in the verification process. All annotators were graduate students in computational linguistics and were familiar with idiomatic expressions and noun-compound interpretation.

\paragraph{Annotation interface.}
For each noun compound instance, annotators were shown:
\begin{enumerate}
    \item the target noun compound text (e.g., \textit{Eye Candy});
    \item the original high-fidelity image from \textsc{AdMIRe}, used as a semantic reference;
    \item four candidate iconographic images generated by our abstraction pipeline.
\end{enumerate}

\paragraph{Task instructions.}
Annotators were instructed to evaluate each candidate along two dimensions:
\begin{enumerate}
    \item \textbf{Semantic preservation:} whether the candidate preserved the intended meaning of the noun compound;
    \item \textbf{Stylistic conformity:} whether the candidate satisfied the required iconographic constraints (flat design, geometric composition, limited palette, minimal background detail).
\end{enumerate}

Annotators selected the best candidate only if at least one image satisfied both criteria. This was \emph{not} a forced-choice task: if all four candidates failed to preserve the intended meaning or violated the stylistic constraints, annotators rejected the entire batch.

\paragraph{Regeneration loop.}
Rejected batches triggered a new generation cycle. Four fresh candidates were produced and submitted for re-annotation under the same protocol. This generation--verification loop continued until one candidate passed the semantic and stylistic checks.

\paragraph{Stopping criterion.}
An instance was finalized only when at least one candidate was judged acceptable under both criteria and selected as the best rendering for that noun compound and target sense.

\paragraph{Statistics.} Across the full dataset (200 test instances $\times$ 5 image categories = 1,000 items), the first-round acceptance rate was 78.4\%. The remaining 21.6\% required one regeneration cycle; fewer than 3\% required two or more cycles. Inter-annotator agreement (Fleiss' $\kappa$) on the accept/reject decision was 0.81, indicating almost perfect agreement.

\section{Qualitative Failure Cases}
\label{app:failure_cases}

We provide representative examples for the two main failure types discussed in Section~\ref{sec:qualitative_failure}. In each case, we compare model behavior under high-fidelity and iconographic conditions and describe the observed error pattern.

\subsection{Failure Type I: Hyper-fidelity Literal Pull}

This failure occurs when visually rich images over-emphasize constituent objects, textures, or scene details, causing the model to prefer a literal interpretation over the intended idiomatic one.

\paragraph{Example 1: \textit{Eye Candy}.}
In the high-fidelity condition, the image contains visually salient object-level cues associated with literal \textit{eyes} and entertainment-related artifacts. Several models anchor on these literal objects and assign higher semantic fit to the literal rendering than to the idiomatic one. Under iconographic abstraction, these distractor cues are suppressed, and stronger generative models show a substantial reduction in signed literal bias.

\paragraph{Example 2: \textit{Night Owl}.}
In the high-fidelity condition, the presence of an owl-like figure and nighttime context attracts the model toward a literal interpretation. After simplification into a schematic representation of the intended idiomatic meaning, generative models reduce this literal preference, although some discriminative models still retain a positive bias toward the literal reading.

\subsection{Failure Type II: Semantic Drift under Partial Abstraction}

This failure occurs when a model no longer commits to a fully literal reading, but still fails to align with the intended idiomatic meaning. Instead, it drifts toward visually or semantically adjacent alternatives.

\paragraph{Example 1: \textit{Paper Tiger}.}
After iconographic abstraction, the model no longer focuses on photorealistic animal detail, but still overweights surface-level cues associated with \textit{paper} or \textit{tiger} independently. As a result, it selects a semantically nearby but incorrect candidate rather than the intended idiomatic depiction.

\paragraph{Example 2: \textit{Cold Shoulder}.}
In this case, abstraction reduces the influence of high-fidelity scene detail, yet the model remains attracted to concrete visual elements associated with bodily posture or temperature. The resulting prediction reflects partial semantic overlap rather than the conventional idiomatic meaning.

\paragraph{Summary.}
Across these cases, the qualitative evidence supports the quantitative results in Table~\ref{tab:main_results}. High-fidelity images tend to amplify literal visual attraction, while iconographic abstraction reduces this pull. However, simplification alone does not fully solve idiomatic grounding: weaker models, especially discriminative encoders, may still drift toward semantically related but incorrect interpretations.
\section{External Validation and Statistical Details}
\label{sec:appendix_external_validation}

\paragraph{Human correlation study.} To validate that $\Delta$ captures a meaningful underlying phenomenon, we conducted a blinded human evaluation. We randomly sampled 100 noun compounds (200 images) from our dataset. Three independent annotators rated the idiomatic alignment of each image on a 1--5 Likert scale, strictly blinded to model outputs, architectures, and automated $\Delta$ scores.

Inter-annotator agreement was substantial (Fleiss' $\kappa = 0.76$). We computed ``Human $\Delta$'' as the absolute difference in mean human ratings between literal and idiomatic images for each NC. Spearman rank correlations between Human $\Delta$ and automated $\Delta$: Discriminative/Cosine ($\rho = 0.64$, $p < 0.001$), Generative/Log-Prob ($\rho = 0.69$, $p < 0.001$), and Proprietary/Confidence-Score ($\rho = 0.73$, $p < 0.001$).

\paragraph{Confidence score vs.\ choice-frequency validation.} GPT-5: $\rho = 0.82$; Claude 4.5 Sonnet: $\rho = 0.79$. Both signals capture related but not identical aspects of preference, confirming the confidence-based $\Delta$ values in Table~\ref{tab:main_results}.

\paragraph{Bootstrap 95\% Confidence Intervals for $\Delta$ of the Mean (10,000 resamples, $n=200$ instances).}

\begin{center}
\small
\begin{tabular}{lcc}
\toprule
\textbf{Model} & \textbf{$\Delta$ AdMIRe [95\% CI]} & \textbf{$\Delta$ DIVA [95\% CI]} \\
\midrule
SigLIP 2        & [0.241, 0.249] & [0.174, 0.182] \\
EVA-CLIP-18B    & [0.257, 0.267] & [0.187, 0.195] \\
MetaCLIP 2      & [0.246, 0.256] & [0.180, 0.188] \\
InternVL3 (78B) & [0.135, 0.141] & [0.087, 0.091] \\
Qwen2.5-VL (32B)& [0.142, 0.148] & [0.093, 0.098] \\
LLaVA-OV (7B)  & [0.172, 0.180] & [0.119, 0.125] \\
GPT-5           & [0.063, 0.067] & [0.020, 0.022] \\
Claude 4.5 Sonnet & [0.070, 0.074] & [0.027, 0.029] \\
\bottomrule
\end{tabular}
\end{center}
\noindent\small\textit{CIs are computed as mean $\pm 1.96 \times \mathrm{SD}/\sqrt{200}$, reflecting the precision of the mean estimate across 200 instances.}

\paragraph{Wilcoxon signed-rank tests.} All eight models show significant $\Delta_{photo} > \Delta_{icon}$ reductions ($p < 0.001$), confirming the iconographic abstraction effect is robust.

\section{Examples}
\label{app:examples}

\begin{figure*}[t]
    \centering
    \setlength{\tabcolsep}{1pt}

    \begin{minipage}{0.02\textwidth}
        \centering
        \rotatebox{90}{\textbf{\footnotesize AdMIRe (High-Fidelity)}}
    \end{minipage}%
    \hfill
    \begin{subfigure}[b]{0.185\textwidth}
        \centering
        \textbf{\footnotesize High-Literal} \par\medskip
        \includegraphics[width=\textwidth]{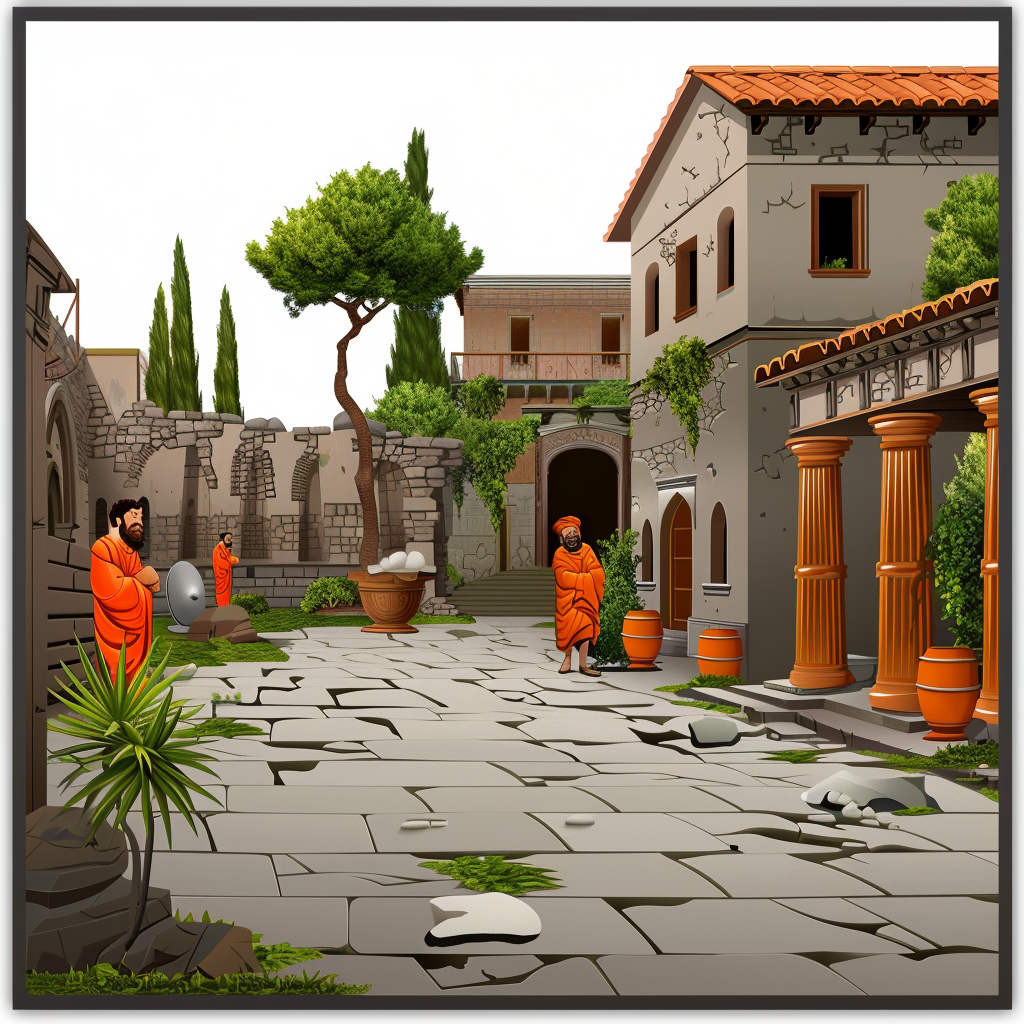}
    \end{subfigure}
    \hfill
    \begin{subfigure}[b]{0.185\textwidth}
        \centering
        \textbf{\footnotesize Weak-Literal} \par\medskip
        \includegraphics[width=\textwidth]{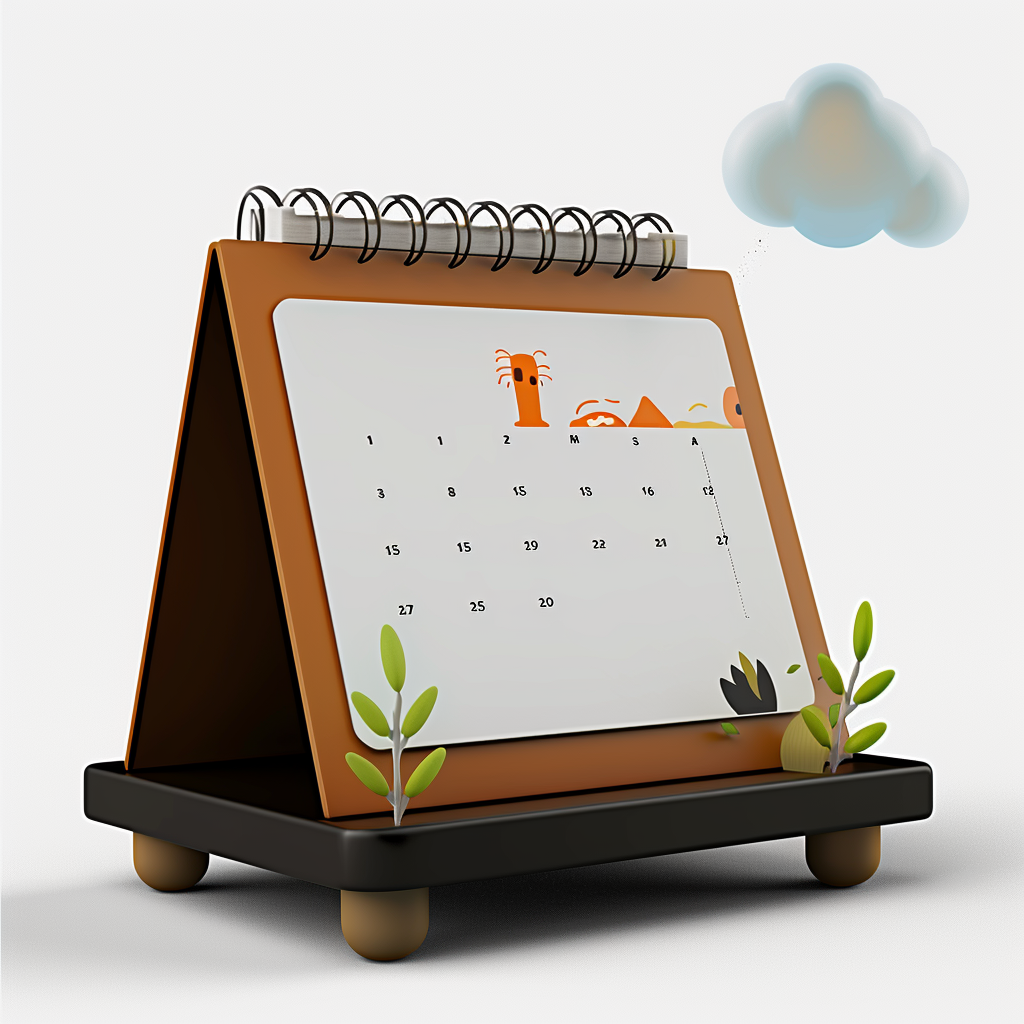}
    \end{subfigure}
    \hfill
    \begin{subfigure}[b]{0.185\textwidth}
        \centering
        \textbf{\footnotesize Weak-Idiomatic} \par\medskip
        \includegraphics[width=\textwidth]{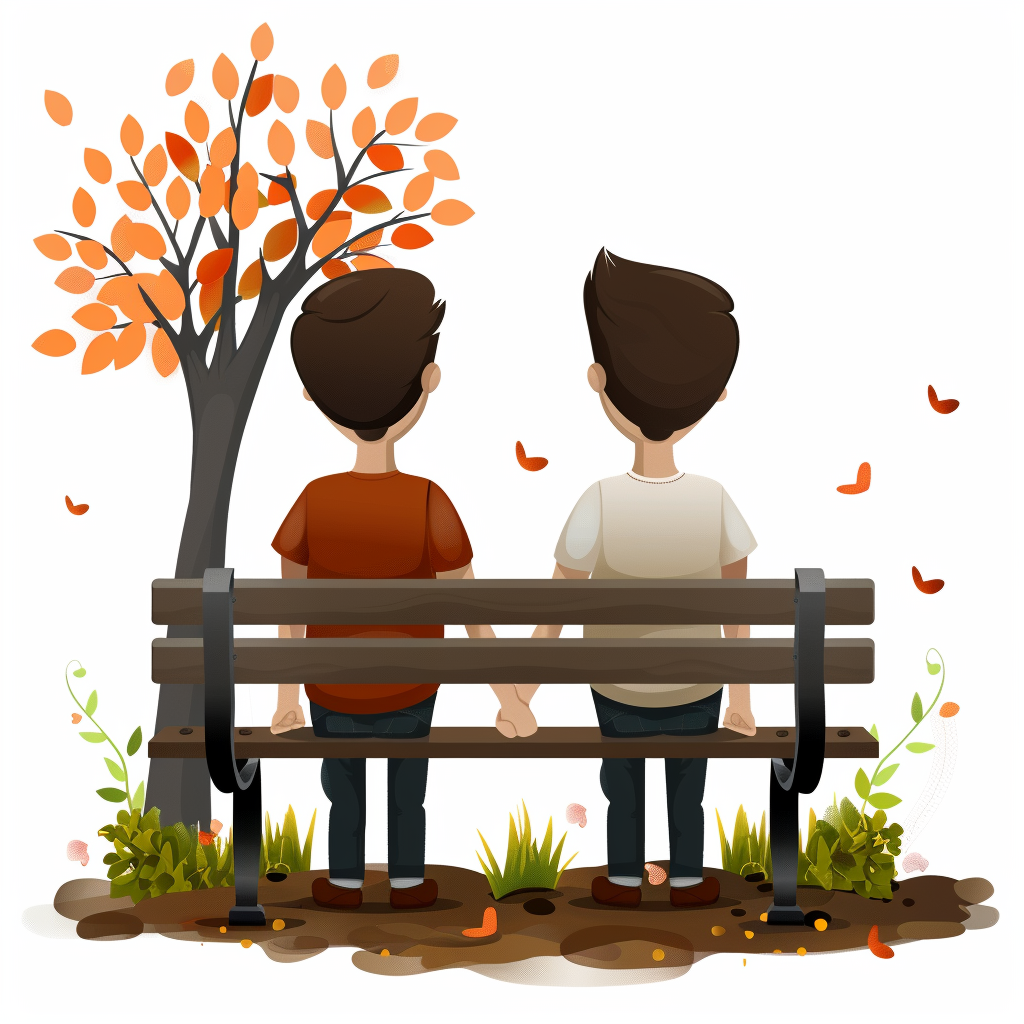}
    \end{subfigure}
    \hfill
    \begin{subfigure}[b]{0.185\textwidth}
        \centering
        \textbf{\footnotesize High-Idiomatic} \par\medskip
        \includegraphics[width=\textwidth]{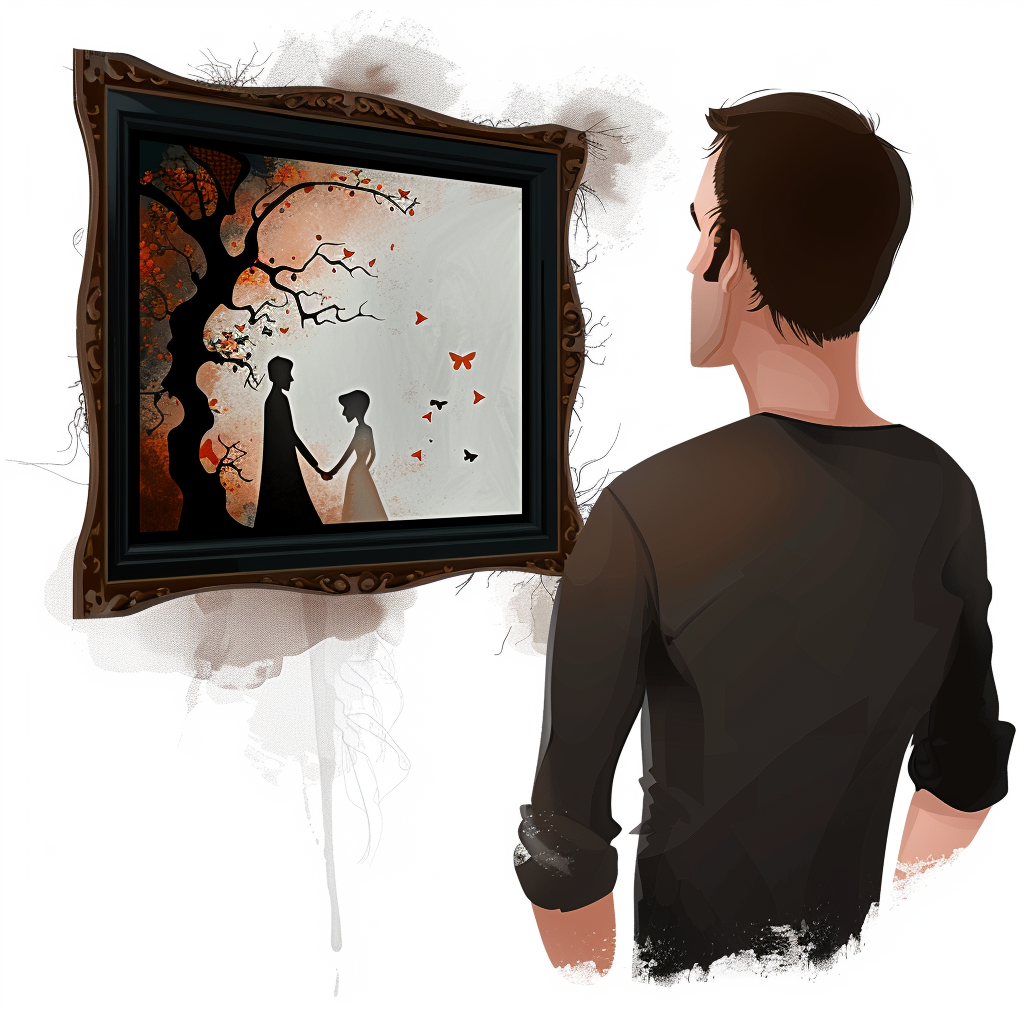}
    \end{subfigure}
    \hfill
    \begin{subfigure}[b]{0.185\textwidth}
        \centering
        \textbf{\footnotesize Distractor} \par\medskip
        \includegraphics[width=\textwidth]{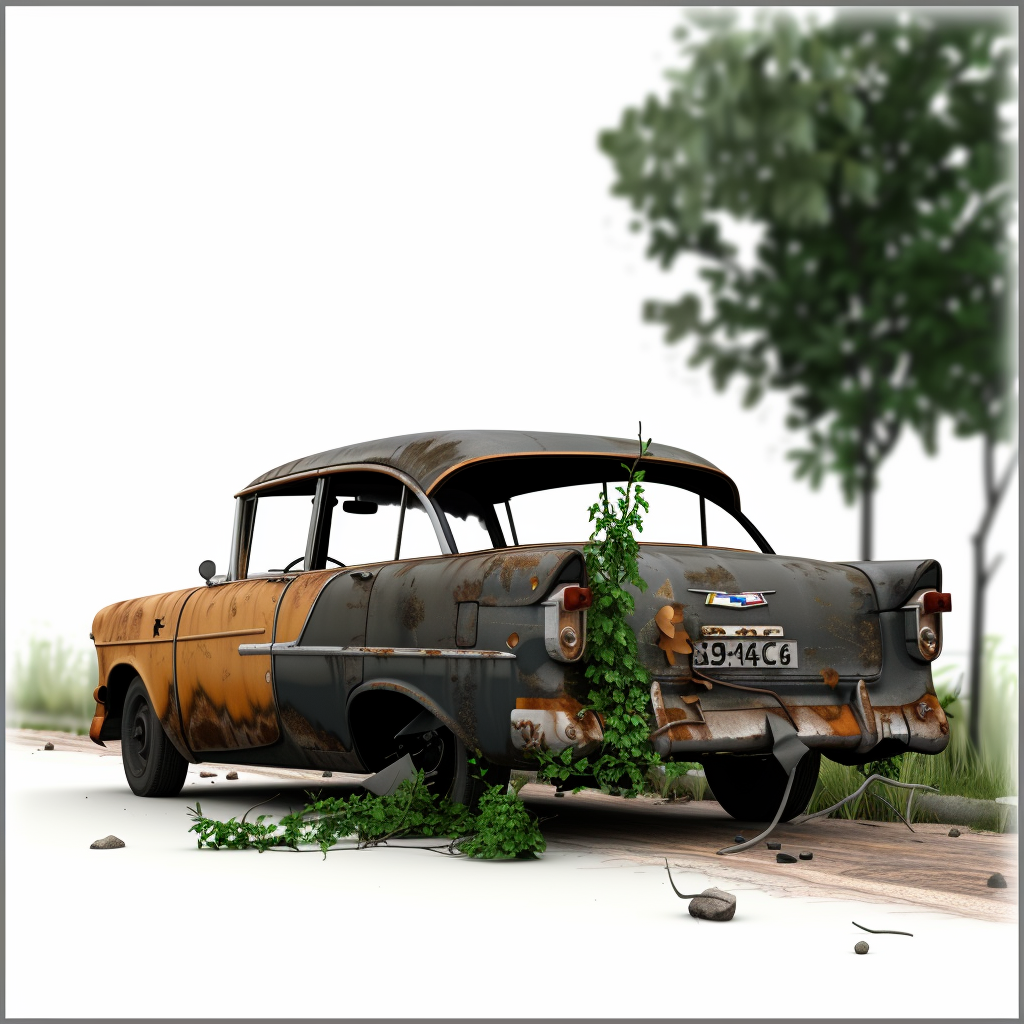}
    \end{subfigure}

    \vspace{0.5em}

    \begin{minipage}{0.02\textwidth}
        \centering
        \rotatebox{90}{\textbf{\footnotesize DIVA (Ours)}}
    \end{minipage}%
    \hfill
    \begin{subfigure}[b]{0.185\textwidth}
        \centering
        \includegraphics[width=\textwidth]{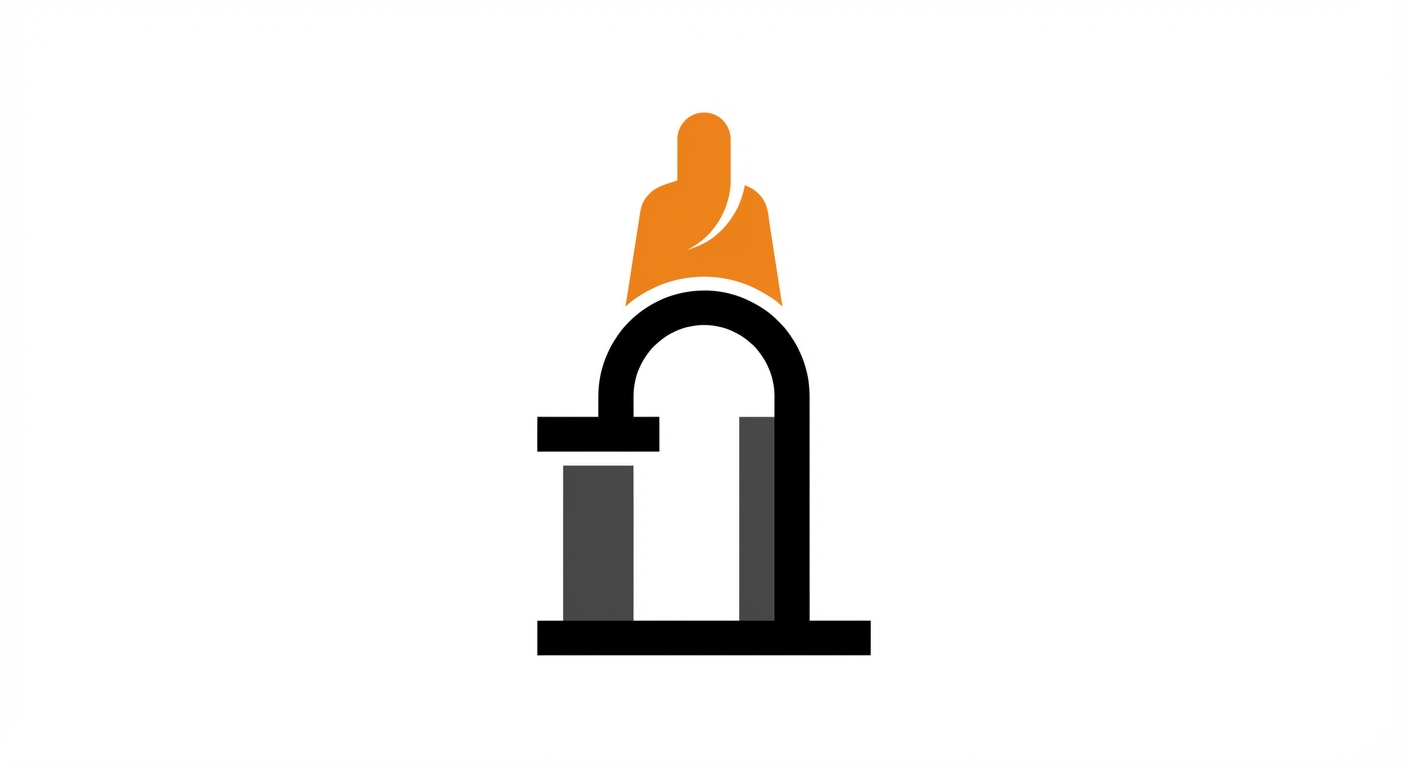}
    \end{subfigure}
    \hfill
    \begin{subfigure}[b]{0.185\textwidth}
        \centering
        \includegraphics[width=\textwidth]{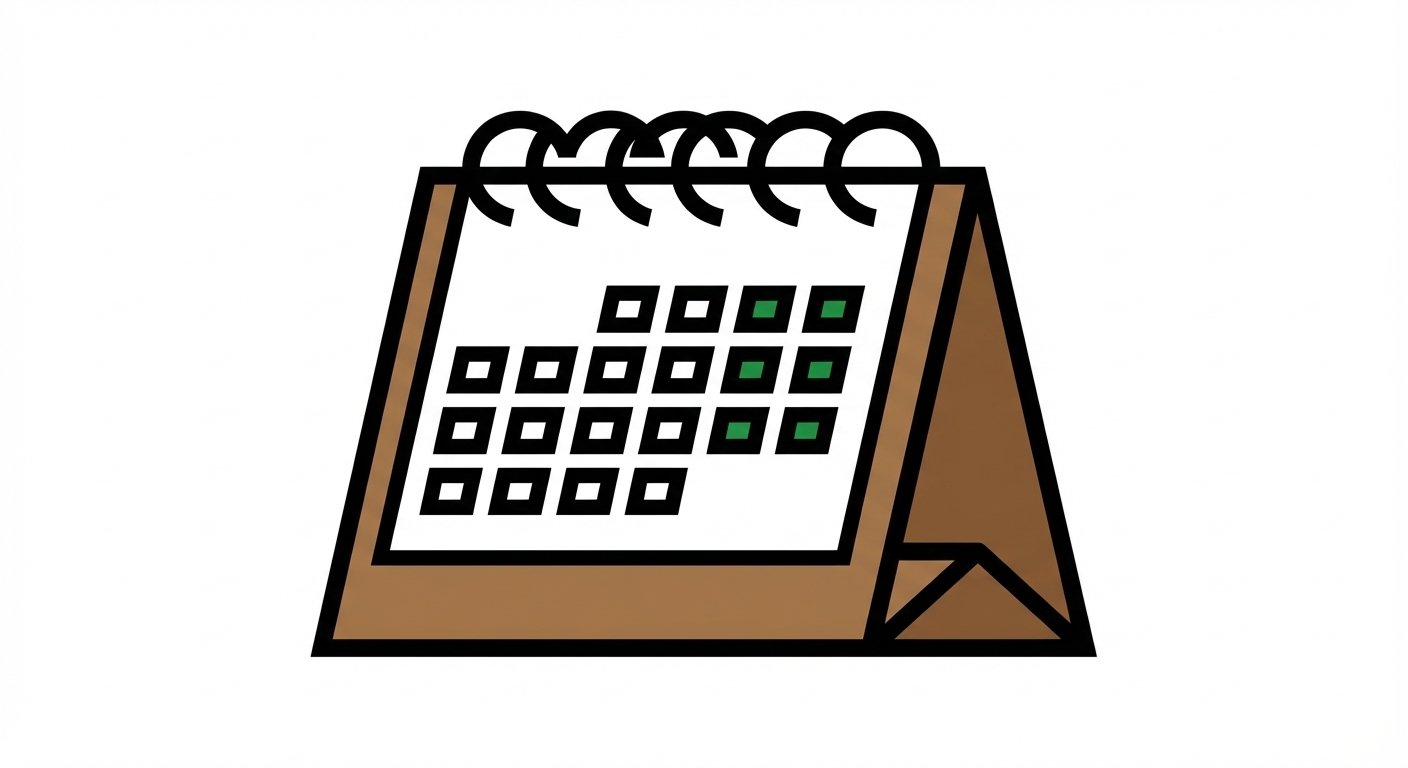}
    \end{subfigure}
    \hfill
    \begin{subfigure}[b]{0.185\textwidth}
        \centering
        \includegraphics[width=\textwidth]{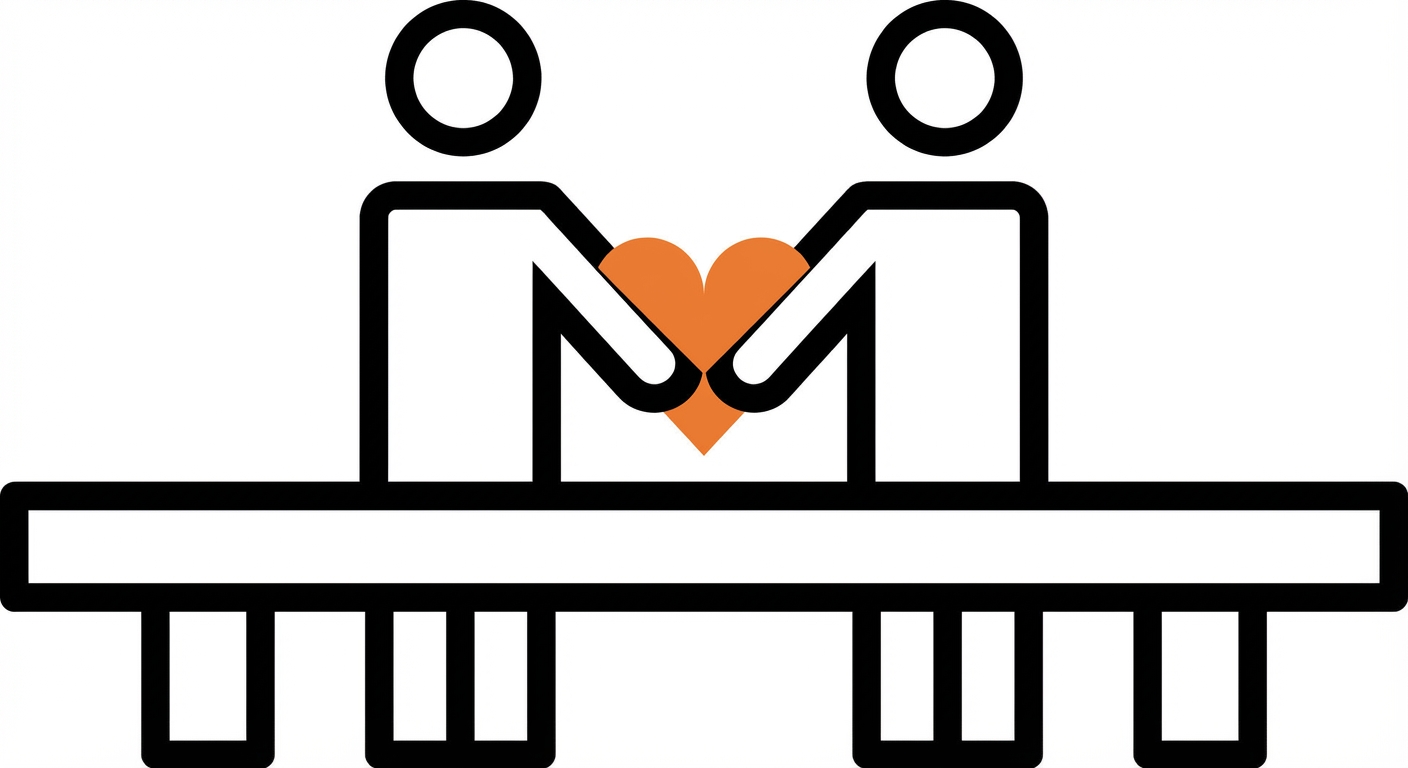}
    \end{subfigure}
    \hfill
    \begin{subfigure}[b]{0.185\textwidth}
        \centering
        \includegraphics[width=\textwidth]{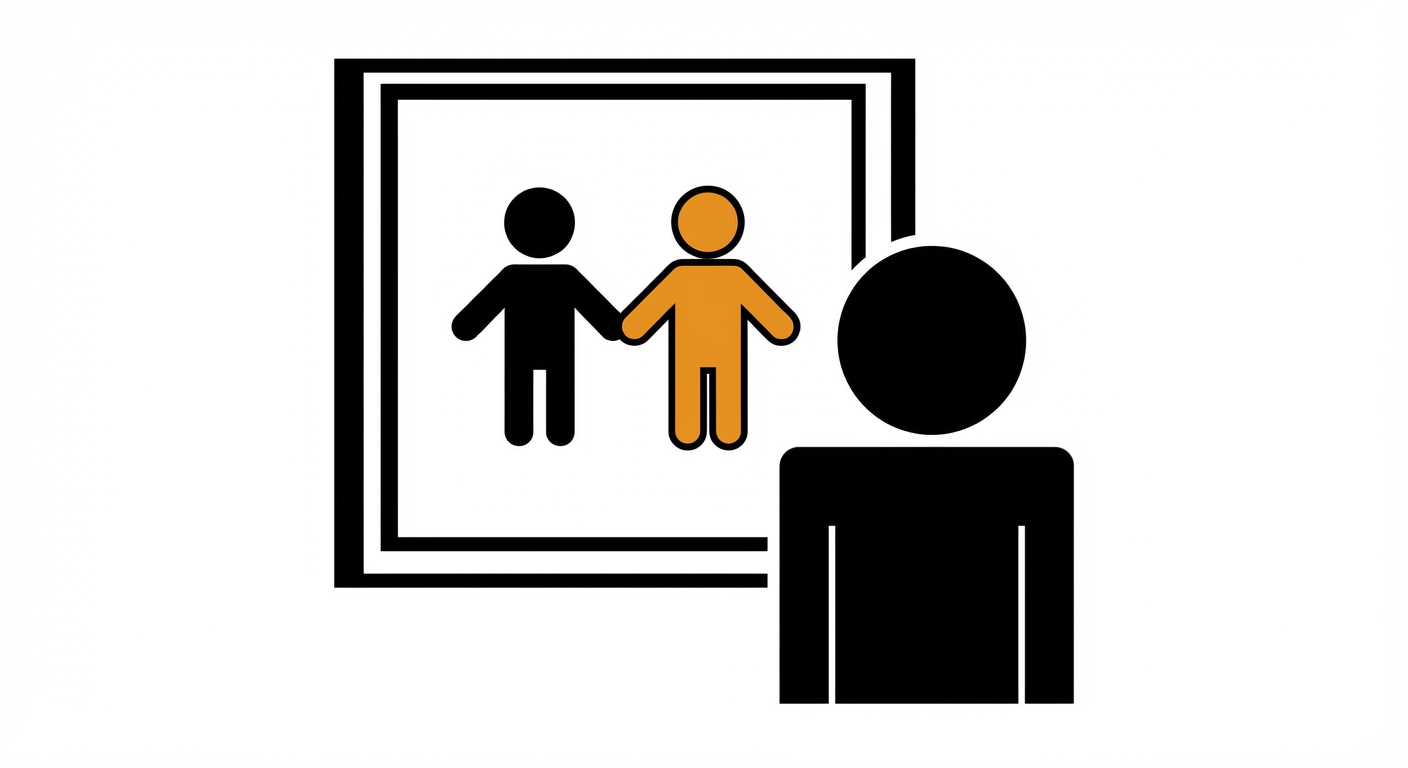}
    \end{subfigure}
    \hfill
    \begin{subfigure}[b]{0.185\textwidth}
        \centering
        \includegraphics[width=\textwidth]{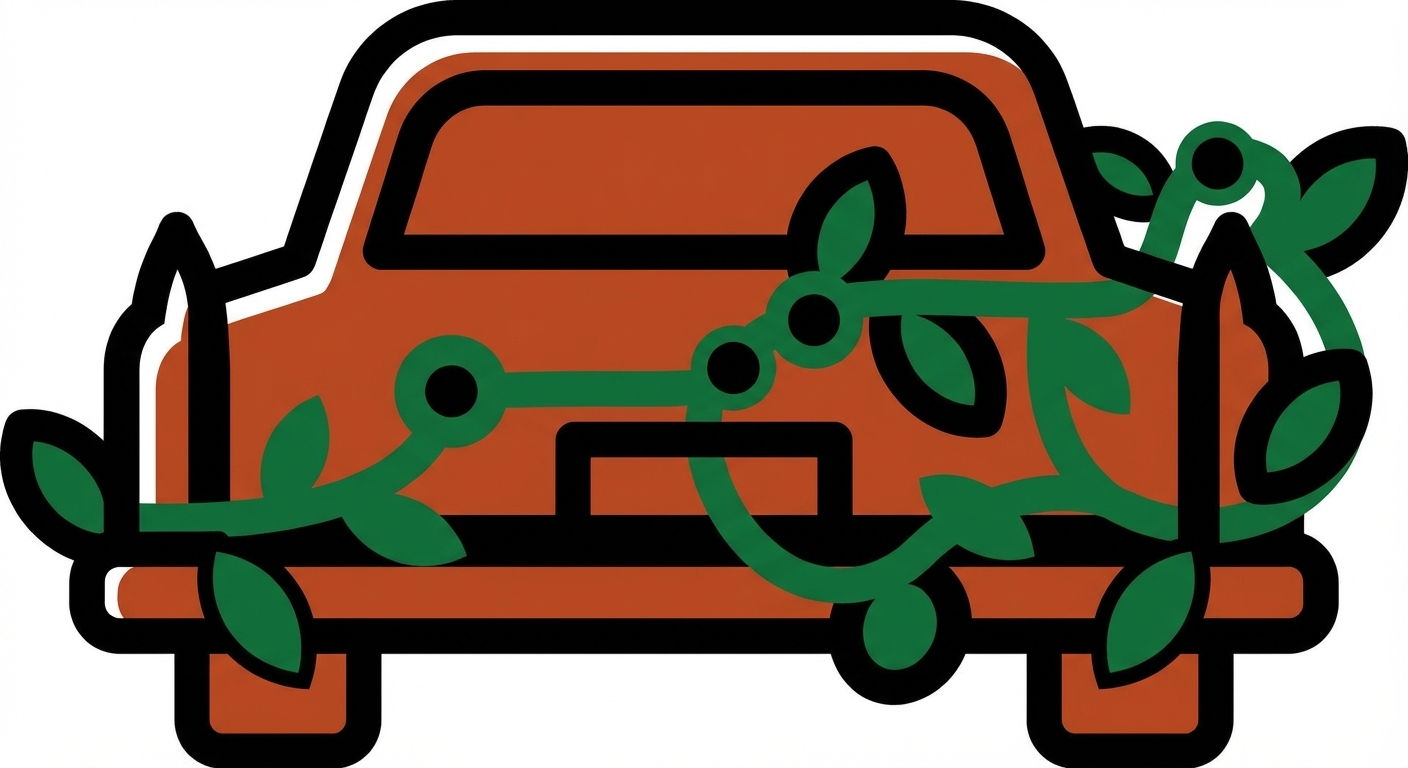}
    \end{subfigure}

    \caption{\textbf{Iconographic Abstraction in Action (AdMIRe vs.\ \textsc{DIVA}).}
    Top Row: The original high-fidelity images from \textsc{AdMIRe}, where high-frequency texture creates ``semiotic noise.''
    Bottom Row: Our corresponding \textsc{DIVA} icons.
    By systematically simplifying the images across the full semantic spectrum (from Literal to Idiomatic), \textsc{DIVA} provides a clean, structure-aware testbed for multimodal reasoning.}
    \label{fig:comparison_panel}
\end{figure*}

\end{document}